\documentclass{article}
\usepackage[utf8]{inputenc}
\usepackage[a4paper, margin=1in]{geometry}
\usepackage{graphicx}
\usepackage[colorlinks=true, allcolors=blue,hypertexnames=false,pagebackref=true]{hyperref}
\usepackage[font=small,labelfont=bf]{caption}

\usepackage{amsmath}
\usepackage{amsfonts}
\usepackage{authblk}

\usepackage{booktabs}

\usepackage{mathabx}

\usepackage{svg}
\usepackage{dsfont} 
\usepackage{multirow} 
\usepackage{array}  
\usepackage{arydshln} 
\usepackage{pbox} 

\usepackage[left]{lineno}
\nolinenumbers
\usepackage{setspace}
\usepackage{makecell}
\usepackage{tabularx}
\newcolumntype{Y}{>{\centering\arraybackslash}X}
\newcolumntype{C}{>{\centering\arraybackslash}p{4.5em}}
\usepackage{float}

\usepackage{changepage}                 
\usepackage{placeins}
\let\Oldsection\section
\renewcommand{\section}{\FloatBarrier\Oldsection}
\let\Oldsubsection\subsection
\renewcommand{\subsection}{\FloatBarrier\Oldsubsection}
\let\Oldsubsubsection\subsubsection
\renewcommand{\subsubsection}{\FloatBarrier\Oldsubsubsection}

\newlength{\offsetpage}
{\end{adjustwidth}}

\def\R{\mathbb{R}}

\newcommand{\norm}[1]{\left\|#1\right\|}

\def\R{\mathbb{R}}

\def\Nc{\mathcal{N}}

\def\argmax{\mathop{\rm arg\,max}\limits}

\def\maxop{\mathop{\rm max}\limits}

\title{Generating Realistic Counterfactuals for Retinal Fundus and OCT Images using Diffusion Models}

\author[1,2,$\mathsection$]{Indu Ilanchezian}
\author[2,3,$\mathsection$]{Valentyn Boreiko}
\author[4]{Laura K\"uhlewein}
\author[1,2]{Ziwei Huang}
\author[5]{Murat Se\c{c}kin Ayhan}
\author[2,3]{Matthias Hein}
\author[1,2*]{Lisa Koch}
\author[1,2*]{Philipp Berens}

\affil[1]{Hertie Institute for AI in Brain Health, University of T\"ubingen, Germany}
\affil[2]{Tübingen AI Center, T\"ubingen, Germany}
\affil[3]{Department of Computer Science, University of Tübingen, T\"ubingen, Germany}
\affil[4]{Center for Ophthalmology, University of T\"ubingen, Germany}
\affil[5]{Institute of Ophthalmology, University College London, London, UK}
\affil[$\mathsection$]{\normalfont{Joint first authors}}
\affil[*]{\normalfont{Senior authors: \texttt{lisa.koch@uni-tuebingen.de}, \texttt{philipp.berens@uni-tuebingen.de}}}

\date{}

\begin{document}

\maketitle

\begin{abstract}
Counterfactual reasoning is often used in clinical settings to explain decisions or weigh alternatives. Therefore, for imaging based specialties such as ophthalmology, it would be beneficial to be able to create counterfactual images, illustrating answers to questions like "If the subject had had diabetic retinopathy, how would the fundus image have looked?". Here, we demonstrate that using a diffusion model in combination with an adversarially robust classifier trained on retinal disease classification tasks enables the generation of highly realistic counterfactuals of retinal fundus images and optical coherence tomography (OCT) B-scans. The key to the realism of counterfactuals is that these classifiers encode salient features indicative for each disease class and can steer the diffusion model to depict disease signs or remove disease-related lesions in a realistic way. In a user study, domain experts also found the counterfactuals generated using our method significantly more realistic than counterfactuals generated from a previous method, and even indistinguishable from real images.  
\end{abstract}

\section{Introduction}

Humans naturally use counterfactual thoughts, deliberations and statements to reason about the causal structure of the world, understand the past and prepare for the future \cite{byrne2016counterfactual}. For example, counterfactuals are used in medicine to explain decisions or weigh alternatives: ``If we had treated the patient with drug X, she might have experienced severe side effects.''\cite{prosperi2020causal}. In a similar way, when medical images are available for diagnosis, it might be useful to create counterfactual images that visualize the answer to the question: "For a given subject who we believe is healthy, how would the imaging data have looked for the same subject to be identified as the diseased class?".  

In ophthalmology, for instance, clinicians regularly use imaging modalities such as retinal fundus photography and Optical Coherence Tomography (OCT)  for diagnosing sight-threatening diseases like Diabetic Retinopathy (DR) and Age-Related Macular Degeneration (AMD). Counterfactual images as described above can be generated from Deep Neural Networks (DNNs) that are trained to detect the presence of these diseases. In this context, counterfactuals are artificially generated images that contain minimal, realistic, meaningful and high-confidence changes to an input image such that the DNN classifier alters its decision to a desired target class \cite{svce_gcpr2022}. For them to look realistic and meaningful, the models used to create them need to have outstanding generative abilities. The resulting images can then also be viewed as explanations of the DNN's decisions as they enable the user of a DNN model to visualise the features that the classifier relies on for detecting the disease \cite{svce_miccai2022,attrinet_midl2023}. 

Previously, different strategies for generating counterfactuals have been proposed \cite{styleGANCEs_2021, svce_gcpr2022, svce_miccai2022, attrinet_midl2023, gifsplanation_midl2021, ageCEs_2023,lesion_cfs_2022, anomaly_cfs_2022}. For example, DNN-based counterfactuals can be generated by iteratively superimposing the input image with the gradients of an adversarially robust classifier, which has more informative gradients than a plain model \cite{svce_gcpr2022,svce_miccai2022, robustprop_neurips2019}. While these so-called sparse counterfactuals show meaningful features, they appear to modify the original image in unexpected and unnatural ways. On fundus images, they cover lesions with unnatural blood vessels in order to generate healthy counterfactuals \cite{svce_miccai2022}. In a similar vein, when StyleGANs are used to generate counterfactual retinal fundus images for a Diabetic Macular Edema classifier \cite{styleGANCEs_2021}, the counterfactuals generated with this procedure begin to show features relevant to target class even before the decision of the classifier changes, despite their highly realistic appearances.  Counterfactuals of OCT scans have also been generated using GANs to study retinal aging, but domain experts were easily able to identify the generated images \cite{ageCEs_2023}, suggesting that they are not sufficiently realistic. Finally, in other medical domains such as brain tumor detection from MRI images and chest X-ray interpretation, counterfactuals based on diffusion models have been used to generate healthy counterfactuals from diseased images \cite{lesion_cfs_2022, anomaly_cfs_2022}, but not for generating images showing a disease from healthy ones. 

Here, we show that we can generate realistic counterfactual generation within the context of retinal disease detection from two ophthalmic imaging modalities, funduscopy and OCT, by relying on deep generative models known as diffusion models \cite{diffusionCEs_2022}. These diffusion models have been shown to outperform GANs in realistic image generation, while also overcoming their drawbacks by producing diverse samples and covering a broad range of the image distribution in tandem with a stable training process \cite{ddpm_2020, guideddiff_neurips2021}. We use classifiers trained to detect several eye diseases from retinal images and then show how to combine these with a generative diffusion model to result in realistic counterfactual retinal images that explain classifiers' decisions in both directions: from healthy to diseased and vice versa. Importantly, we show that domain specialists -- ophthalmologists and AI experts -- view the resulting images as realistic when probed in an odd-one-out task. This indicates that our methods generates images that fulfill the criteria for counterfactual images to be used in medical reasoning as outlined above.

\section{Methods}

We first describe the ophthalmic imaging datasets used in this study and then review the relevant methods for the generation of counterfactuals for such images. Lastly, we describe our design of a user study in order to evaluate the clinical relevance of counterfactuals. 

\subsection{Datasets}

We used retinal image data sets from two common ophthalmic imaging modalities: (1) color fundus photography (CFP) and (2) Optical Coherence Tomography (OCT). 

Fundus images were obtained from EyePacs Inc. through a Diabetic Retinopathy (DR) screening program\footnote{\url{https://www.eyepacs.com/blog/over-750-000-patients-screened}}. Initially, this collection contained over $180,000$ retinal fundus images from over $42,000$ subjects along with meta data such as age, sex, race and blood pressure. Image quality was indicated as "Insufficient for Full Interpretation", "Adequate", "Good" or "Excellent" per image as annotated by Eyepacs Inc. Some DR labels were missing. 
We used "Good" and "Excellent" quality images with DR labels only, resulting in $92,745$ retinal fundus images from $27,926$ participants. Then, we created training, validation and test splits subject-wise (see Table \ref{tab:datasets}). The training set was augmented with $789$ images from the Benitez data set \cite{Benitez_2021} and $1842$ images from the FGADR data set \cite{fgadr_itmi2021} in order to strengthen the representation of diseased samples for the diffusion models. All images were cropped to square dimensions of $224 \times 224$ pixels using a circle fitting procedure (\url{https://github.com/berenslab/fundus_circle_cropping/tree/v0.1.0},\cite{mueller_fundus_circle_cropping_2023}).

For OCT B-scans, we used a data set consisting of a total of $108,309$ images belonging to one of four categories \cite{kermany_2018} : normal, choroidal neovascularization (CNV), drusen and Diabetic Macular Edema (DME) (Table \ref{tab:datasets}). In order to obtain a square center crop including the macular region, we used only the images with size $496 \times 512$ and $496 \times 768$ ($96,441$ scans). We created training, validation and test splits again subject-wise with $75\%$ subjects in training, $15\%$ in validation and $10\%$ in the test set, respectively (see Table \ref{tab:datasets}). 

\begin{table}[t]
	
	\centering
    \caption{Summary of the retinal image collections used for model development and evaluation.}
    \label{tab:datasets}
    \begin{tabular}{@{\extracolsep{\fill}}p{0.3cm}p{1.75cm}p{2.0cm}p{2.0cm}p{2.5cm}p{2.5cm}p{2.5cm}}
    \toprule
        &   &   &   & Training   &  Validation &   Test \\
        \midrule
        \multirow{21}{\textwidth}{\rotatebox[origin=c]{90}{CFP}} & \multirow{7}{\textwidth}{EyePacs}   &   subjects & & 15,827 & 5,324 & 6,775 \\
            \cmidrule{3-7}
            &                   &   images &  all & 46,921  &  15,658  &  30,166 \\
            &                   &          &   healthy  & 38,502 & 12,748   &  24,627\\  
            &                   &          &   mild  & 3,244 &  1,163 &  2,378\\ 
            &                   &          &   moderate  & 4,695 & 1,572  & 2,907\\
            &                   &          &   severe  & 238 &  121  & 127 \\
            &                   &          &   proliferative  & 242 & 54  & 127 \\
            \cmidrule{2-7}
            & \multirow{5}{\textwidth}{Benitez}      
            &   images &  all & 789  &  -  & - \\
            &          &          & healthy & 94 & - &  -\\  
            &          &          &   mild  & 6 & -  & -\\ 
            &          &          &   moderate & 102 &  - & -\\
            &          &          &   severe+  & 587 & -  & -\\
            \cmidrule{2-7}
            & \multirow{5}{\textwidth}{FGADR}      
            &   images &  all & 1,842    &  -  & - \\
            &     &     &  healthy & 101 & - & - \\ 
            &     &     &   mild  & 212 & -  & -\\ 
            &     &     &   moderate  & 595 &  - & -\\
            &     &     &   severe+  & 934 & - & -\\
        \midrule
        \multirow{6}{\textwidth}{\rotatebox[origin=c]{90}{OCT}} & \multirow{6}{\textwidth}{Kermany}   &   subjects  &  & 3558  & 712 & 474  \\
        \cmidrule{3-7}
        &   &   images  &  all & 71,231  & 14,714  &  10,496 \\
        &   &           &  normal & 34,340 &  6,813 &  4,464 \\
        &   &           &  CNV & 23,133 &  5,091 &   3,738    \\
        &   &           &  drusen & 5,393 & 1,221 &  1,288 \\
        &   &           &  DME & 8,365 &  1,589  &   1,006  \\
    \bottomrule
		\end{tabular}
\end{table}

\subsection{Generating realistic counterfactual retinal images}

As mentioned above, we define visual counterfactuals as minimal, realistic and high-confidence changes to an image $x_0$ by which a classifier's prediction can be altered to a desired target class \cite{svce_gcpr2022}. They show what features are important for the classifier to change the decision to a particular class, and hence provide insights into what is learned by the classifier. Since the generative capabilities of a classifier are typically limited and it cannot by itself generate realistic counterfactuals, we rely on a diffusion model \cite{guideddiff_neurips2021} to achieve realism. In order to generate counterfactuals, the reverse diffusion process is modified such that classifier gradients contribute to this process and guide the diffusion model towards producing counterfactuals in the desired class \cite{dvce_neurips2022}. 

We will first discuss diffusion models in Section\,\ref{sec_diffusion} and the various types of classifiers used here in Section\,\ref{sec_adversarial} before we introduce Diffusion Visual Counterfactuals (DVCs) in Section\,\ref{sec_dvces}. Then, in Section\,\ref{sec_svces}, we briefly describe Sparse Visual Counterfactuals (SVCs) as a baseline method for counterfactual generation with retinal fundus images. Finally, in Section\,\ref{sec_user_study}, we present the details of a user study conducted with clinicians and AI experts to evaluate the realism of generated counterfactuals.

\subsection{Diffusion Models}\label{sec_diffusion}

Diffusion models  are generative image models that yield high-quality and realistic images as a result of two processes \cite{ddpm_2020,guideddiff_neurips2021}: forward diffusion and reverse diffusion. 
Forward diffusion is 
a Markov chain 
that gradually adds Gaussian noise to a starting image $x_{0}$: 
\begin{equation}
    q(x_{t}|x_{t-1}) = \Nc(x_{t}; \sqrt{1-\beta_{t}}x_{t-1}, \beta_{t}\mathbb{I}) ~,
\end{equation}
where $t \in \{1, \dots, T\}$, $\beta_{t}$ denotes a noise schedule such that $q(x_{T}|x_{0}) \approx \Nc(x_{T}; 0, \mathbb{I})$. Given $x_{0}$, 
the noisy images at any time step $t$ can be also expressed in closed form: 
\begin{equation}
    x_{t} = \sqrt{\overline{\alpha}_{t}}x_{0} + \sqrt{1 - \overline{\alpha}_{t}}\epsilon \, , \epsilon \sim \Nc(0, \mathbb{I}), 
    \label{diffusionkernel}
\end{equation}
where $\overline{\alpha}_{t} = \prod_{s=1}^{t} (1 - \beta_{s})$.

Then, in the reverse diffusion process, the posterior $q(x_{t-1}|x_{t}, x_{0})$, when conditioned on $x_0$, can be estimated using the Bayes Theorem \cite{sohl_icml2015}. The unconditioned posterior $q(x_{t-1}|x_{t})$ is, however, intractable and has to be approximated by a parameterized distribution $p_{\theta}(x_{t-1}|x_t)$:
\begin{equation}
    p_{\theta}(x_{t-1}|x_t) = \Nc(x_{t-1}; \mu_{\theta}(x_{t}, t), \Sigma_{\theta}(x_{t}, t)). 
    \label{prior}
\end{equation} 

The mean and diagonal covariance of this distribution are predicted by DNNs denoted by $\mu_{\theta}(x_{t}, t)$ and $\Sigma_{\theta}(x_{t}, t)$, respectively. Briefly, these models are trained by optimizing a simplified loss function derived from the Variational Lower Bound (VLB) of the negative log likelihood $- \log p_\theta(x_0)$. The simplification involves learning the residual noise $\epsilon_{\theta}(x_t,t)$ at each time step and then expressing the mean $\mu_{\theta}(x_{t}, t)$ in terms of $\epsilon_{\theta}(x_t,t)$. $\Sigma_{\theta}(x_{t}, t)$ is modeled as an interpolation between $\beta_t$ and $\tilde{\beta}_t = \frac{1 - \overline{\alpha}_{t-1}}{1 - \overline{\alpha}_t} \beta_t$ using a vector $\mathbf{v}$ that is output by the DNN. For further details about the loss functions and training, see \cite{guideddiff_neurips2021}. Using 
$\mu_{\theta}(x_{t}, t)$ and $\Sigma_{\theta}(x_{t}, t)$, one can generate images from the data distribution $p(x)$ by starting with a sample from the standard normal distribution and iteratively reconstructing less noisy images at previous time steps from the current noisy image at time step $t$. 

For both fundus and OCT data sets, we trained the diffusion model $p_{\theta}$ for $300,000$ minibatch iterations unconditionally with $1,000$ time steps and a linear noise schedule for the diffusion process. The diagonal covariance $\Sigma_{\theta}$ are also learned by the model during training. For the fundus data set, classes are balanced by oversampling the diseased classes to have a equal representation as that of the healthy class. 

\subsection{Plain and adversarially robust classifiers}\label{sec_adversarial}
 
The sampling procedure from Eqn.\,\eqref{prior} results in 
unconditional samples from $p(x)$ whereas counterfactuals must belong to a specified target class, thus requiring conditional sampling from $p(x|y)$. Classifiers can be used to drive diffusion models towards producing realistic images that belong to a desired class \cite{guideddiff_neurips2021, glide_icml2022, BD_cvpr2022}. More specifically, the gradients of a classifier with respect to the image shift the mean of the reverse transitions (Eqn.\,\eqref{prior}) to guide the diffusion model in the right direction. Nevertheless, a plain classifier does not have gradients that are perceptually aligned with the features of a particular class and could result in counterfactuals that look visually similar to the original image when the changes are constrained to be minimal. In contrast, the gradients of adversarially robust models have strong generative properties and are more effective in guiding diffusion models towards generating meaningful features for a target class \cite{robustprop_neurips2019,dvce_neurips2022} despite being subjected to a constraint for producing minimal changes.  

This property of adversarially robust models can be attributed to their training procedures which expose them to adversarial attacks. Consider a $K$-class classifier $f_{\phi}$ with parameters $\phi$, logits $f_{\phi}(x) \in \mathbb{R}^K$ and output probabilities $p_{\phi}(c|x) \in [0,1]^K$ where $x \in \R^d$ is the input to the classifier and $c \in \{1, \dots, K\}$. A targeted adversarial attack adds imperceptible perturbations to a starting image $x_0$ which changes the decision of the classifier from the correct class to a target class $k$. More precisely, an $\ell_p$ targeted adversarial attack for $f_{\phi}$ at $x_0$ produces a sample $x$, such that
     
\begin{equation}
 \argmax_{c \in \{1, \dots, K\}} f_{\phi}(x)_c = k, \quad x \in [0,1]^{d} \cap B_p(x_0, \varepsilon). 
    \label{adversarial_sample}
\end{equation}
where $B_p(x_0, \varepsilon) := \{\hat{x} \in \R^d | \norm{x_0-\hat{x}}_p \leq \varepsilon \}$ is an $\ell_p$ ball around the original image $x_0$ with radius $\varepsilon$.
One usually maximizes a surrogate loss $L$ for this:
\begin{equation}
    \argmax_{x \in [0,1]^{d} \cap B_p(x_{0}, \varepsilon)} L(f_{\phi}(x), k).
    \label{adversarial_example}
\end{equation}

To defend the classifier $f_{\phi}$ empirically against such attacks, one can perform adversarial training. A well-known and commonly used algorithm for this is TRADES \cite{trades_icml2019}. Its loss function incorporates a term for the adversarial examples in addition to the standard cross-entropy loss:
\begin{equation}
   \frac{1}{n}\sum_{i=1}^n \big[ -\log\big(p_{\phi}(y_i| x_i)\big) + 
    \beta \maxop_{x \in B_2(x_i, \varepsilon)} D_{KL} \big( p_{\phi}(\cdot | x) \,||\, p_{\phi}(\cdot | x_i) \big)\big],
    \label{eq_TRADES}
\end{equation}
where $D_{KL}$ denotes the Kullback-Leibler divergence
and $\beta$ 
controls the trade-off between adversarial and plain training schemes. In our experiments, we set $\beta$ to $6$ \cite{trades_beta_iccv2021, trades_icml2019}. This process results in a classifier $f_\psi$ which is robust to adversarial perturbations. Plain classifiers $f_\phi$, on the other hand, are not robust to adversarial attacks and corresponds to training with only the cross-entropy loss i.e. $\beta$ is $0$. 

For retinal fundus images, we trained both plain and adversarially robust classifiers in binary and multi-class settings. In the multi-class setting, the task is a 5-way classification among the classes ``healthy'', ``mild'', ``moderate'', ``severe'' and ``proliferative''. In the binary setting, disease onset is considered from the ``moderate'' class, hence, ``healthy'' and ``mild'' are grouped into the normal category and the other classes to the diseased category. For OCT scans, we trained both plain and adversarially robust classifiers in the multi-class setting to classify among the classes ``healthy'', ``choroidal neovascularization (CNV)'', ``drusen'' and ``diabetic macular edema (DME)''. 

All plain and robust classifiers were ResNet-50 models trained for $100$ epochs with an SGD optimizer with a learning rate of $0.01$ and a cosine learning rate schedule. The fundus plain classifiers were initialized with weights from ImageNet pre-trained models and the fundus robust classifier with weights from a robustly pre-trained ImageNet model \cite{MadryRobust_2019}. All OCT classifiers were initialized with random weights. We used the cross-entropy (CE) loss as objective function for the plain model and the TRADES loss \cite{trades_icml2019} with $\varepsilon=0.01$ for the fundus robust classifiers and $\varepsilon=0.5$ for the OCT classifier. For both cases, we used $p=2$.  

\subsection{Diffusion Visual Counterfactuals}\label{sec_dvces}
 Here, we describe how to produce realistic Diffusion Visual Counterfactuals (DVCs). Following \cite{dvce_neurips2022}, we combined an unconditionally trained diffusion model $p_\theta$ as described in Section~\ref{sec_diffusion} with an independently trained classifier $f_{\phi}$ (see Section\,\ref{sec_adversarial}) so that the diffusion model can generate class-conditional samples.

 \begin{figure}[h!]
\includegraphics[width=\textwidth]{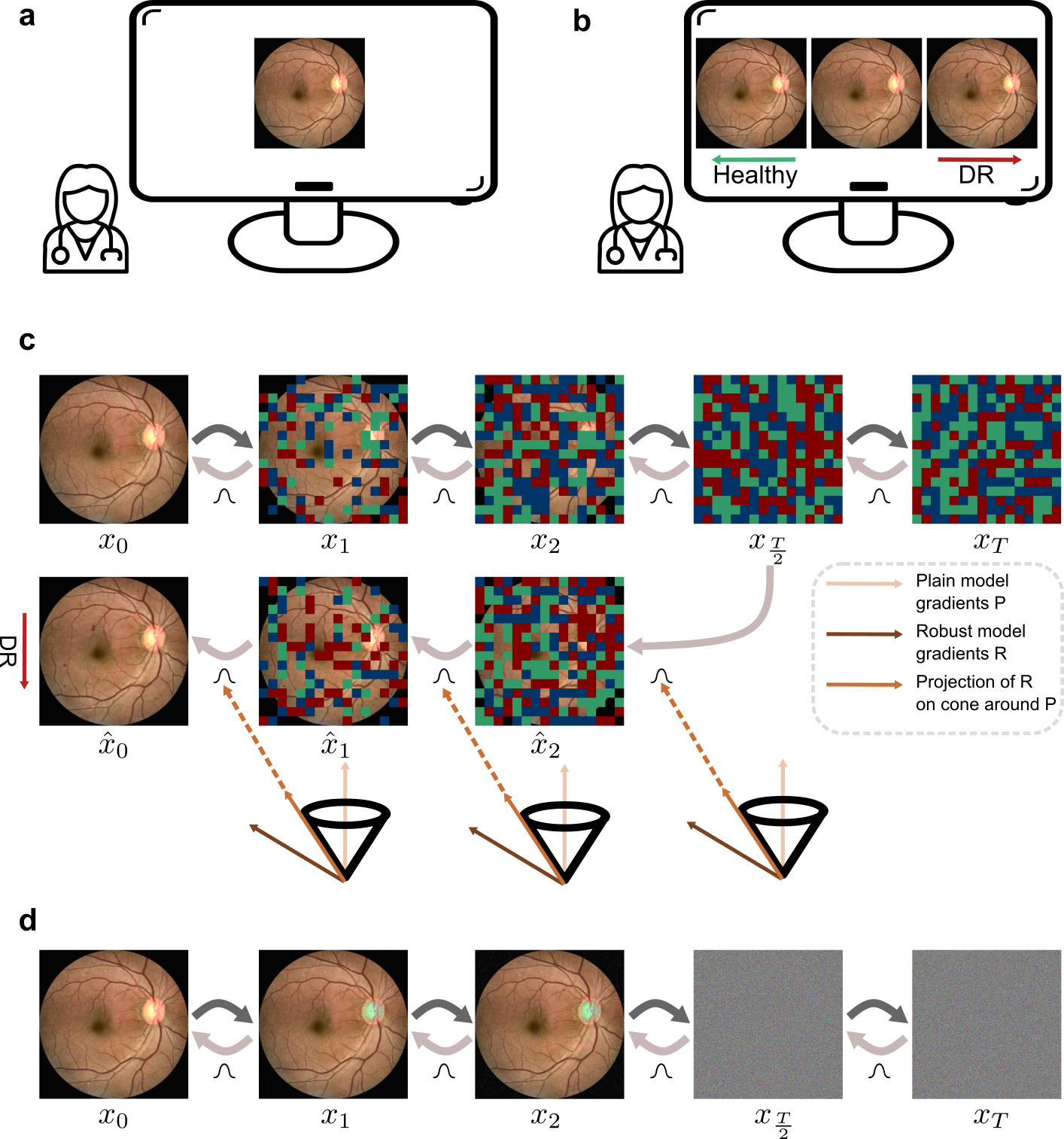}
\caption{\textbf{a.} Original retinal fundus image, \textbf{b.} Visualization of counterfactuals with the healthy counterfactual on the left, DR counterfactual on the right and original image in middle, \textbf{c.} Method to generate diffusion counterfactuals. Top shows the forward and reverse diffusion for an original image $x_0$. Bottom shows generation of a DR DVC starting from the $\frac{T}{2}^{th}$ time step. The mean of distributions in reverse diffusion is shifted using projected gradients (shown in dark orange) of an adversarially robust classifier (shown in brown) on a cone around the gradients of a plain classifier (shown in light orange), \textbf{d.} Images from the actual forward diffusion corresponding to the time steps shown in c.}
\label{fig1}
\end{figure}

In general, classifiers used in conjunction with diffusion models for conditional sampling are noise-aware, i.e., they are trained on noisy images which occur at various time steps in the diffusion process \cite{guideddiff_neurips2021}. Consequently, the input to these classifiers includes the time step $t \in \{1,...,T\}$ at which the image $x_t \in \R^d$ occurred. 
Since our classifiers are not noise-aware or time-dependent, that is to say the inputs are only the images $x \in \R^d$, we use an estimation of $x_0$ from a given noisy sample $x_t$ as input to the classifier. We denote this estimated $x_0$ by $x_{0, dn}(x_t, t)$ and following Eqn.\,\eqref{diffusionkernel} it can be expressed as:   
\begin{equation}
    x_{0,dn}(x_t, t) \rightarrow \frac{x_t}{\sqrt{\bar{\alpha_{t}}}} - \frac{\sqrt{1-\bar{\alpha_{t}}}\epsilon_{\theta}(x_t,t)}{\sqrt{\bar{\alpha_{t}}}},
    \label{eq_denoised}
\end{equation}
where $\epsilon_{\theta}(x_t,t)$ can be calculated as a function of the mean $\mu_{\theta}(x_t,t)$ \cite{guideddiff_neurips2021}.
The reverse process transitions of a diffusion model guided by an external classifier which is not noise-aware are given by: 
\begin{equation}
    p_{\theta, \phi}(x_{t-1} | x_{t}, y) = Zp_{\theta}(x_{t-1}|x_{t})p_{\phi}(y|x_{0,dn}({x_t, t}))
    \label{eq_condsample}
\end{equation}
where $Z$ is a normalization constant. 
Exact sampling from this distribution is intractable, however, it can be approximated by a Gaussian distribution in a way similar to the unconditional reverse transitions (Eqn.\,\eqref{prior}) but with shifted mean \cite{guideddiff_neurips2021}\cite{dvce_neurips2022}.
\begin{linenomath}
\begin{align}
    & p_{\theta,\phi}(x_{t-1} | x_t, y) = \Nc(\mu_t, \Sigma_{\theta}(x_t, t)), \\
    & \mu_t = \mu_{\theta}(x_t, t) + \Sigma_{\theta}(x_t, t) \nabla_{x_t} \log p_{\phi} (y|x_{0,dn}(x_{t}, t))
    \label{eq_sampling}
\end{align}
\end{linenomath}
The shift in the mean depends on the gradients of the external classifier which guide the diffusion model to generate images in a specified target class (Fig.\,\ref{fig1}). This, however, does not ensure that the generated image will stay close to the original image $x_0$ in pixel space, which is one of the qualifying factors for realistic visual counterfactuals. 
Therefore, to obtain a counterfactual from \( p(x|y) \) that remains structurally close to the original image, \( x_0 \), we find it beneficial to add a distance regularization term to Eqn.\,\eqref{eq_sampling}. To maintain consistent parameters \( \lambda_c, \lambda_d \) across different images, an adaptive parameterization, as discussed in \cite{dvce_neurips2022}, is important. This adaptation changes the mean of the transition kernel to:
\begin{linenomath}
\begin{align}
    & \mu_t = \mu_{\theta}(x_t, t) + \Sigma_{\theta}(x_t, t) \norm{\mu_{\theta}(x_t, t)}_2 \Gamma_{\textrm{DVC}}, \\
    & \Gamma_{\textrm{DVC}} = \lambda_c \frac{\nabla_{x_t} \log p_{\phi} (y|x_{0,dn}(x_t, t))}{\norm{\nabla_{x_t} \log p_{\phi} (y|x_{0,dn}(x_t, t))}_2} - \lambda_d \frac{\nabla_{x_t} d(x_0, x_{0,dn}(x_t, t))}{\norm{\nabla_{x_t} d(x_0, x_{0,dn}(x_t, t))}_2}
    \label{update}
\end{align}
\end{linenomath}
 As a further measure to avoid generating images that deviate too much from the original, we start the reverse of the diffusion process from the noisy image at step $\frac{T}{2}$ instead of the completely distorted version of the image at the last step $T$ \cite{dvce_neurips2022}(Fig.\,\ref{fig1}). 
 
In Eqn.\,\eqref{update}, the plain model $f_{\phi}$ can also be replaced by the adversarially robust model $f_{\psi}$. While adversarially robust models have stronger generative properties, they suffer from a considerable drop in accuracy compared to plain models. Hence, it would be advantageous to explain a plain model with better performance while also utilizing the stronger gradients of the robust model. To achieve this, we project the gradients of the adversarially robust model $\nabla_{x_t} \log p_\psi (y|x_{0,dn}(x_t, t))$ onto a cone centered around the gradients of the plain model $\nabla_{x_t} \log p_{\phi} (y|x_{0,dn}(x_t, t))$ (Fig.\,\ref{fig1}). This procedure is called cone projection \cite{dvce_neurips2022} 
 and it is done by changing $\Gamma_{\textrm{DVC}}$ in Eqn.~\eqref{update} to
 \begin{linenomath}
\begin{align}
\Gamma_{\textrm{DVC}} = \lambda_c \frac{\Gamma_{\textrm{cone}}}{\norm{\Gamma_{\textrm{cone}}}_2} - \lambda_d \frac{\nabla_{x_t} d(x_0, x_{0,dn}(x_t, t))}{\norm{\nabla_{x_t} d(x_0, x_{0, dn}(x_t, t))}_2},
\label{update_cone}
\end{align}
\end{linenomath}
where 
\begin{equation}
    \Gamma_{\textrm{cone}} = P_{\textrm{cone}(\alpha, \nabla_{x_t} \log p_{\phi} (y|x_{0,dn}(x_t, t))}\big[\nabla_{x_t} \log p_{\psi} (y|x_{0,dn}(x_t, t))\big],
\end{equation}
$\lambda_c$ and $\lambda_d$ are positive constants and $\alpha$ is the angle of the cone, which we set to $30^\circ$ following \cite{dvce_neurips2022}.

Our code was based on  \url{https://github.com/valentyn1boreiko/DVCEs} and will be available at \url{https://github.com/berenslab/retinal_image_counterfactuals} .

\subsection{Prior work: Sparse Visual Counterfactuals}\label{sec_svces}

Previous studies on generating retinal counterfactuals either use StyleGANs \cite{styleGANCEs_2021} or adversarially robust classifiers \cite{svce_miccai2022}. While the StyleGAN approach is closer to our approach as it uses a generative model, the code or model information is not adequately provided for reproducing the results presented. Hence, for comparison, we used a previously suggested method for generating Sparse Visual Counterfactuals (SVCs) requiring an adversarially robust classifier \cite{svce_gcpr2022} or at least an ensemble of plain and adversarially robust classifiers \cite{svce_miccai2022}. Sparse counterfactuals are computationally similar to adversarial examples but conceptually different from them due to the fact that sparse counterfactuals show meaningful changes that are relevant to the target class instead of the imperceptible noise added to original examples. 

For generating sparse counterfactuals, we used the log probability of the target class as a surrogate loss function (Eqn.\,\eqref{adversarial_example}):
\begin{equation}
    L(f_{\psi}(x), y) = -\log p_{f_{\psi}}(y|x)
    \label{VCE}
\end{equation}

The sparsity and degree of realism of the generated counterfactuals can be controlled by changing the norm used for defining the constrained set $B_p(x_0, \varepsilon)$. A norm of $\ell_{4}$ was shown to generate the most realistic counterfactuals among the various norms \cite{svce_miccai2022}. Since closed-form projections are not possible for $\ell_{4}$ norm, the Auto-Frank-Wolfe (AFW) algorithm \cite{svce_gcpr2022} was used to solve the optimization and generate sparse counterfactuals \cite{svce_miccai2022}. 

The main drawback of sparse counterfactuals is that visual inspection showed that healthy counterfactuals from the DR class using retinal fundus images covered up the lesions on the fundus image with artificial looking blood vessels in a previous study \cite{svce_miccai2022}. Although this achieved the effect of removing the lesions to make the image look healthy, these changes do not appear realistic (e.g. see Fig.\,2, second row in \cite{svce_miccai2022}). A more realistic change would have been to cover up the lesions with the background colors instead of adding artificial structures.

\begin{table}[t]
\centering
 \caption{Evaluation of plain and robust classifiers in terms of standard and balanced accuracy}
        \label{tab:accuracies}
        \begin{tabular}{@{\extracolsep{4pt}}lCCCCCC@{}}
        \hline
        & \multicolumn{2}{c}{Binary fundus}
        & \multicolumn{2}{c}{5-class fundus} 
        & \multicolumn{2}{c}{OCT} \\
        \cline{2-3} \cline{4-5} \cline{6-7}
        & \multicolumn{1}{C}{acc.} & \multicolumn{1}{C}{bal. acc.} & \multicolumn{1}{C}{acc.}  & \multicolumn{1}{C}{Quad. $\kappa$} &
        \multicolumn{1}{C}{acc.}  & \multicolumn{1}{C}{bal. acc} \\ 
        \hline
        Plain & $92.39$ & $80.67$  & $86.65$ & $0.67$ & $96.35$ & $95.87$ \\
        Robust
        & $90.03$ & $74.35$ & $83.69$  & $0.51$ & $95.03$ & $93.29$ \\
        \hline
        \end{tabular}
\end{table}

\subsection{User study}\label{sec_user_study}
To evaluate the realism of the generated counterfactuals, we performed a user study with AI experts as well as trained ophthalmologists. We built a web-based image evaluator based on the Python web framework Django (v. 4.2.1) with a PostgreSQL (v. 15.3) backend database, available at \url{https://github.com/berenslab/retimgtools/tree/v.1.0.0}. On the front-end, we used custom JavaScript to modify various presentation parameters (e.g. hiding the images after a certain number of seconds) . 

Seven ophthalmologists who had a clinical experience of $2, 4, 5, 9, 9, 10$ or $14$ years participated in the study (including author LaK). In addition, $4$ AI experts working on applying deep learning for clinical tasks in ophthalmology took part and provided their input (including authors PB and LiK). 

All participants were given a three-way odd-one-out task where they had to identify the generated counterfactual among three images. This task design is recommended for this type of study as it is highly sensitive for detecting the odd-one-out category \cite{men_against_machines_2017, sensitivity_user_study_2017}. Each trial thus consisted of two real images from the data sets and one counterfactual generated by a model. Images were displayed for a maximum of $20$ seconds and then hidden. All $3$ images in any question belonged to the same class. For example, for a question showing DR images, we show two real DR images and one generated counterfactual with DR as the target class. The latter is generated from an image which is labeled as healthy in the data set and classified as healthy by the classifier.

For retinal fundus images, a total of $80$ trials were performed with a randomly chosen set of $40$ questions showing sparse counterfactuals and the remaining half showing diffusion counterfactuals as the generated image. Within each group, $50\%$ questions belonged to the healthy class and the rest to DR. For OCT scans, on the other hand, only diffusion counterfactuals were shown as the generated images in all $80$ questions as study time was a limiting factor with four disease categories. Questions were equally split across the four disease categories with $20$ questions for each class. Similar to the fundus scenario, OCT counterfactuals for questions belonging to the healthy category are generated from any of the three disease classes and vice-versa. 

Ethical approval for the study was obtained from the ethics commission at the University Clinic, Tübingen (Ref No. 250/2023BO2). Statistical analysis was performed using R. 

\section{Results}

Our goal was to show that the counterfactuals based on diffusion models guided by the gradients of robust classifiers can generate minimal, meaningful and high-confidence changes to an input image such that the DNN classifier alters its decision to a desired target class and that domain experts view the resulting images as realistic. To this end, we first report the result of a user study with domain experts in order to evaluate the realism of our counterfactuals generated with the chosen parameters (Section~\ref{sec_dvces}). We then go into the technical factors necessary for achieving this result, establish that robust classifiers are indeed necessary  and illustrate the effect of regularization strength on the generated images. Finally, we demonstrate multi-class counterfactuals and counterfactuals for retinal OCT scans, for which we also evaluate the realism in a user study.

\subsection{Fundus diffusion counterfactuals are realistic}
We trained binary plain and robust DNN classifiers for DR based on fundus images with high accuracy on a large and diverse fundus image dataset (Table \ref{tab:accuracies}). For details of the dataset, see Table \ref{tab:datasets}; for details of training procedure, see Section~\ref{sec_adversarial}). As expected, the robust classifier had lower accuracy than the plain one. In addition, we also trained a diffusion model on the same dataset augmented by additional datasets to add more diseased examples (for details of the training procedure, see Section~\ref{sec_diffusion}). We used the diffusion model with cone projected gradients in order to generate realistic diffusion counterfactuals such that if an image showed signs of DR, the counterfactual could either remove these signs (``healthy diffusion counterfactual'') or reinforce them (``DR diffusion counterfactual''). Likewise, the diffusion counterfactual could either add signs of DR to a healthy original image or strengthen its healthy appearance. Thus, the model was able to generate images that illustrate what the fundus image of a patient might have looked like, had he or she been more or less progressed in their disease (the definition of a counterfactual). We compared the diffusion counterfactual method to the previously published sparse counterfactuals method \cite{svce_miccai2022}. 

\begin{figure}[t!]
\includegraphics[width=\textwidth]{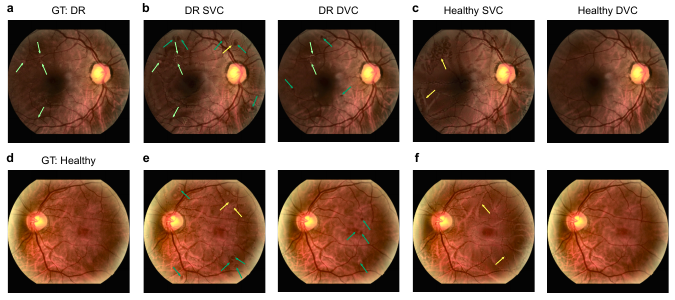}
\caption{DVCs show clinically meaningful changes and appear more realistic than SVCs. \textbf{a.} Image with DR and classifier confidence $p_{\phi}(\textrm{DR}) = 0.99$. \textbf{b.} DR SVC (left) and DVC (right) with $p_{\phi}(\textrm{DR}) = 1.00$ for both images. \textbf{c.} Healthy SVC (left) with $p_{\phi}(\textrm{healthy}) = 1.00$ and healthy DVC (right) with $p_{\phi}(\textrm{healthy}) = 0.99$.  DVCs show realistically emphasized lesions (light green arrow) and new lesions (dark green arrow). DVC shows more realistic removal of disease related lesions whereas SVCs introduce artifacts (yellow arrow). \textbf{d.-f.}, as \textbf{a.-c.}, but for a healthy fundus image $p_{\phi}(\textrm{healthy}) = 0.90$.  DR SVC: $p_{\phi}(\textrm{DR}) = 0.98$; DR DVC: $p_{\phi}(\textrm{DR}) = 1.00$; Healthy SVC: $p_{\phi}(\textrm{healthy}) = 1.00$; healthy DVC: $p_{\phi}(\textrm{healthy}) = 0.99$. All SVCs were generated with $\ell_4$ norm and $\epsilon=0.3$. DVCs were generated with $\ell_2$ norm and regularization strength $\lambda=0.5$. } 
\label{fig2}
\end{figure}

\begin{figure}[t!]
\includegraphics[width=\textwidth]{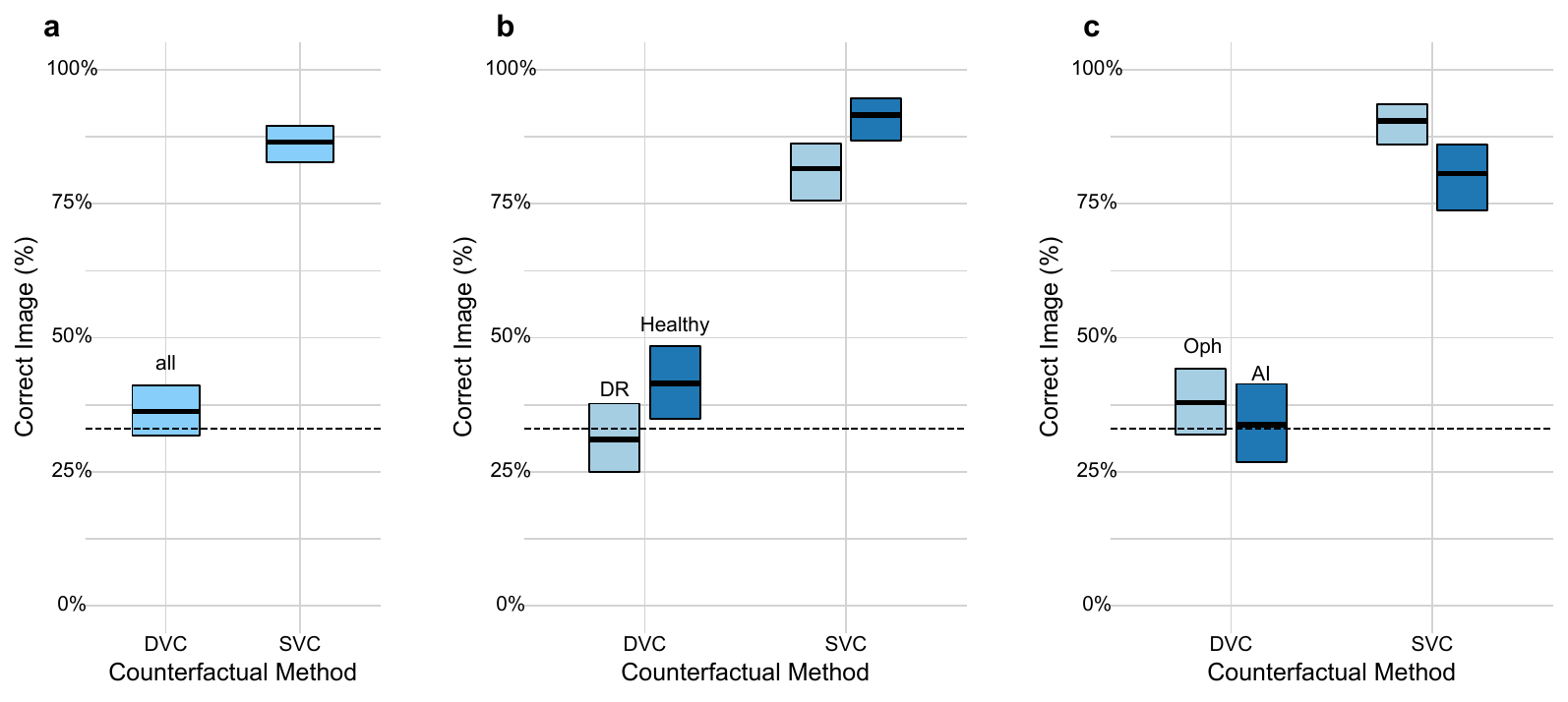}
\caption{User study of realism of generated DVCs. We asked $n=4$ AI experts and $n=6$ ophthalmologists to identify a counterfactual in a odd-one-out task with three images (two real and one counterfactual). \textbf{a.} Overall fraction of correctly identified counterfactuals with binomial 95\%-CI. Baseline at $33\%$ chance level (dashed line). \textbf{b.} As in \textbf{a.} for the healthy and DR classes. \textbf{c.} As in \textbf{a.} for ophthalmologists and AI experts.}
\label{fig3}
\end{figure}

We found that the diffusion model generated visually realistic counterfactual fundus images from either DR (Fig.\,\ref{fig2}\,\textbf{a}) or healthy starting images (Fig.\,\ref{fig2}\,\textbf{d}). For example, a DR diffusion counterfactual generated from a DR fundus images enhanced the existing lesions and added new lesions (Fig.\,\ref{fig2}\,\textbf{b} right panel). On the other hand, a DR diffusion counterfactual generated from a healthy image produced diverse lesions including images regions that resembled microaneurysms, haemmorhages and exudates (Fig.\,\ref{fig2}\,\textbf{e} right panel). Further, the structural details in the retina including the blood vessels, macula and optic disc were largely preserved on the diffusion counterfactuals of any given subject's fundus image. In comparison, baseline sparse counterfactuals appeared more artificial (left panels in Fig.\,\ref{fig2}\,\textbf{b,c,e,f}). Sparse counterfactuals introduced artifacts such as waves around lesions in DR counterfactuals  (Fig.\,\ref{fig2}\,\textbf{b,e}) and lines in healthy counterfactuals(Fig.\,\ref{fig2}\,\textbf{c,f}). For more examples, see Appendix\,\ref{app2}.

\begin{table}[t!]
\centering
 \caption{Generalized Linear Model to assess the influence of factors in Fig. \ref{fig3}. $n=800$}
\begin{tabular}{cccc} 
 \hline
 Predictor & Odds Ratio & CI & p-value \\ 
 \hline
 SVC vs. DVC & 12.03 & 8.45 – 17.40 & $\lll 0.0001$ \\ 
 healthy vs. DR & 1.82 & 1.30 – 2.57 & 0.0005 \\ 
 Ophthalmologist vs. AI researcher & 0.66 & 0.47 – 0.94 & 0.0197 \\ 
 \hline
 \label{tab:fundus}
\end{tabular}
\end{table}

To assess whether the generated diffusion counterfactuals are realistic, we performed a user study with four AI specialists, who worked with ophthalmological data on a regular basis, and six ophthalmologists with different levels of experience (see Sec. \ref{sec_user_study}). In a three-way odd-one-out task, we asked to identify the image likely to have been generated by an AI model. The shown images included both healthy and DR diffusion and sparse counterfactuals. Interestingly, all participants found it challenging to distinguish diffusion counterfactuals from real fundus images whereas they easily spotted the sparse counterfactuals (Fig.\,\ref{fig3}). In fact, across all images, participants showed a close to chance level ($33\%$) performance for diffusion counterfactuals as opposed to a significantly better performance than chance level for sparse counterfactuals (Fig.\,\ref{fig3}\,\textbf{a}, DVC vs. SVC: 36.3\% [31.7\% -- 41.1\%] correct, 95\% CI), confirmed by statistical analysis ($p \lll 0.0001$, see Table \ref{tab:fundus}). 

We further analyzed if DR or healthy could be more easily identified as artificial. We found that participants could identify healthy diffusion counterfactuals more easily compared to DR diffusion counterfactuals (Fig.\,\ref{fig3}\,\textbf{b}), potentially because diffusion models appear to smooth the image during removal of lesions and sometimes fail to remove all traces of lesions ($p=0.0005$, see Table \ref{tab:fundus}). Finally, we studied whether trained ophthalmologists were more likely to identify diffusion counterfactuals than AI specialists. Interestingly, we found that this difference was not major, with all ophthalmologists independent of experience levels being close to chance level, at a similar level as AI specialists (Fig.\,\ref{fig3}\,\textbf{c}). Ophthalmologists detected sparse counterfactuals at an average rate of $90.4\%$, significantly better than AI specialists who detected the same at an average rate of $80.6\%$ ($p=0.0197$, see Table\,\ref{tab:fundus}).

In summary, we found that our diffusion counterfactual model can generate realistic looking fundus images from both healthy and DR images, emphasizing or removing signs of the disease. We showed that the images generated by our new model are almost impossible to detect even for highly trained experts, in contrast to images created by previous techniques.  

\subsection{Realistic counterfactual examples require robust classifiers}

\begin{figure}[t]
\includegraphics[width=\textwidth]{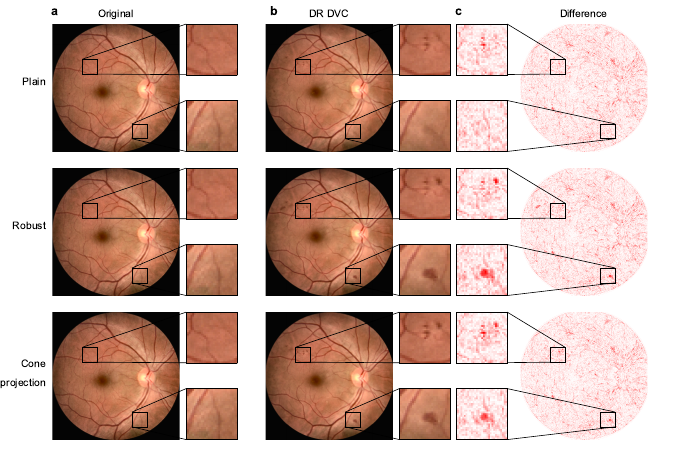}
\caption{Comparison of DVCs generated using the plain model (top row), robust model (middle row) and cone projection of an adversarially robust model onto a plain model (bottom row). \textbf{a.} A DR fundus image with $p_{\phi}(\textrm{DR}) = 1.00$ with a zoom in on patches with lesions. \textbf{b.} DR DVCs for the image from \textbf{a.} for the three different models. \textbf{c.} Difference maps between original DR image and the DR DVC show robust and cone projection models produce more realistic changes than the plain model} 
\label{fig4}
\end{figure}

Now that we established that we are able to generate realistic looking counterfactuals for fundus images, we explore the technical ingredients necessary to achieve this. First, as discussed in Sec. \ref{sec_adversarial}, the gradients of the output a standard classifier with respect to the image often do not represent meaningful changes, but rather lead to the generation of adversarial examples that fool the classifier but are imperceptible for humans \cite{adv_iclr2014}. In fact, for our diffusion model, the gradients of a "plain" classifier often were not strong enough to provide guidance towards the target class and hence, the resulting counterfactuals looked quite similar to the original image (Fig.\,\ref{fig4}\,\textbf{a-c}, top row). This effect was more prominent in DR diffusion counterfactuals generated from healthy images, where the plain classifier's gradients induced hardly noticeable lesions, compared to healthy counterfactuals generated from DR images, where the diffusion model removed lesions even when guided by the plain classifier (compare Fig.\,\ref{fig4} top row to Appendix.\,\ref{app3}). 

In contrast, the gradients of the output of a robust classifier with respect to the image (see Sec. \ref{sec_adversarial}) supported the generation of high quality DR diffusion counterfactuals, with clearly visible and highly realistic lesions (Fig.\,\ref{fig4}\,\textbf{a-c}, middle row). As discussed, the robust classifier, however, traded robustness against accuracy, leading to a drop in performance (see Table \ref{tab:accuracies}). To obtain high quality diffusion counterfactuals while maintaining high classification accuracy, we combined the plain and the robust classifier gradients using cone projection (see Sec. \ref{sec_adversarial}). Here, the gradients of the robust classifier are projected onto a cone around the gradients of the plain classifier. In this case, the generated DR diffusion counterfactuals were almost as good for the robust classifier alone ((Fig.\,\ref{fig4}\,\textbf{a-c}, bottom row and Appendix\,\ref{app4}), while maintaining a high balanced accuracy (Table\,\ref{tab:accuracies}). Therefore, our final model evaluated in the user study above used cone projection for generating realistic diffusion counterfactuals. In the sections that follow, all diffusion counterfactuals are generated using the cone projection method. 

\subsection{Influence of regularisation strength on diffusion counterfactuals}

\begin{figure}[t]
\includegraphics[width=\textwidth]{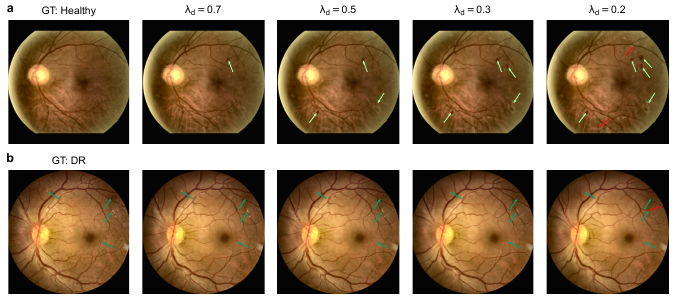}
\caption{Effect of tuning the regularization strength $\lambda_d$ on generated DVCs. Decreasing $\lambda_d$ allows for more changes on the original image. \textbf{a.} We start with a healthy image and generate DR DVCs with decreasing $\lambda_d$. More lesions are generated as $\lambda_d$ decreases (light green arrows). \textbf{b.} We start with a DR image and generate healthy DVCs with different $\lambda_d$ values. Some traces of the lesions were still visible for $\lambda_d = \{0.7, 0.5\}$ while they were completely removed for $\lambda_d = 0.2$ (dark green arrows) at the cost of some changes to the vessel structure (red arrows). While a higher $\lambda_d=0.7$ is sufficient to generate the minimum number of lesions required to convert a healthy fundus to DR, it is not sufficient to remove all lesions on a DR image to convert it to a healthy fundus.} 
\label{fig5}
\end{figure}

Next, we explored the effect of the regularization strength $\lambda_{d}$ (Eqn.\,\eqref{update_cone}), which constrained the distance of the generated diffusion counterfactual from the original image. This parameter controls the extent of changes appearing on the diffusion counterfactuals compared to the original image, with high values indicating that the generated image remains closer to the original. Without regularization, the diffusion model is not constrained to keep the generated image close to the original image and guided by the gradients from the classifier/ cone projection it can generate any image belonging to the target class without necessarily preserving the background of the original image including color and vessel structure. 

We systematically observed the pattern of changes in both DR and healthy diffusion counterfactuals as we lowered the parameter $\lambda_d$. In general, for both DR and healthy diffusion counterfactuals, more changes were visible on images as $\lambda_d$ was decreased. For DR counterfactuals, the size, number and sharpness of lesions increased with decreasing regularization strength. Thus, with a strong regularization of $\lambda_d=\{0.7,0.5\}$, fewer lesions appeared on DR diffusion counterfactuals which were relatively smaller and in some cases not too sharp and distinct (Fig.\,\ref{fig5}\,\textbf{a}). As the strength was decreased to $0.3$, more lesions were generated and their sharpness increased compared to the ones generated with higher regularization values. For all these values of $\lambda_d$, almost all counterfactuals were labeled as DR (see Table \ref{tab:conversion}). At $0.2$, the diffusion model had the freedom to make several modifications to the original image and it added various bright and large lesions while also modifying the blood vessel structure to a larger extent compared to the other regularization values.

\begin{table}
    \centering
    \begin{tabular}{|c|c|c|}
        \hline
        $\lambda_d$ &  Healthy $\rightarrow$ DR & DR $\rightarrow$ healthy\\
         \hline
         0.7 & 4.1\% & 27.0\% \\
         0.5 & 1.4\% & 14.0\%\\
         0.3 & 0\% & 6.0\%\\
         \hline
    \end{tabular}
    \caption{Fraction of images that do not change the class label depending on the choice of regularization parameter $\lambda_d$. For this analysis, 40 images were chosen from each of the five classes, such that there were 80 images for the ``healthy to DR" direction and 120 for ``DR to healthy". For 73/80 and 100/120, class labels were correctly predicted for the original image. Then we evaluated the class label of the corresponding counterfactual.}
    \label{tab:conversion}
\end{table}

We repeated the above for healthy counterfactuals. Notably, we found that lesions were generally removed well in mild and moderate DR already at larger regularization strengths of $\lambda_d=\{0.7,0.5\}$, but some traces of lesions were still visible with for severe, proliferative and a few moderate cases (Fig.\,\ref{fig5}\,\textbf{b}). As regularization was decreased further to $0.3$, the sharpness and clarity of those lesions decreased. At $\lambda_d=0.2$, the lesions were completely removed, however, the blood vessel structure was also heavily altered. For extreme cases of DR, where the entire retina was affected such as in a few proliferate cases, even a regularization of $0.2$ was not sufficient to remove all the lesions (Appendix\,\ref{app5}\,\textbf{a-b}). Furthermore, with $\lambda_d=0.7$, a third of healthy diffusion counterfactuals generated from DR fundus images did not change the prediction of the DNN classifier to the healthy class. In contrast, for $\lambda_d=0.5$ and $\lambda_d=0.3$, 14\% and 6\% did not convert to healthy class, respectively (Table \ref{tab:conversion}). As we evaluated the classifier at a "referable DR" scenario, a healthy diffusion counterfactual may contain small traces of lesions as seen in mild DR fundus images even when the prediction of the classifier changes to healthy. For more examples of diffusion counterfactuals with varying $\lambda_d$, see Appendix\,\ref{app6}. 

Taken together, we found that using values of $\lambda_d$ such as $0.2$ or lower resulted in larger changes to the original image than necessary for conversion to the target class, producing large changes to the vessel pattern. While high $\lambda_d$ such as $0.7$ and higher was sufficient for the DR diffusion counterfactuals to show minimal features required to convert healthy fundus to DR, the same did not hold for healthy diffusion counterfactuals generated from DR fundus images (Table \ref{tab:conversion}). Therefore,  we chose a regularization value of $0.5$ where we could qualitatively observe the minimal changes on the image necessary to alter the decision and confidences of the classifier in both directions while maintaining image structure close to the original (although within a range of $\lambda_d=0.3-0.5$, this is frankly a qualitative judgment).  

\subsection{Diffusion counterfactuals for the multiclass DR grading task}

\begin{figure}[t]
\includegraphics[width=\textwidth]{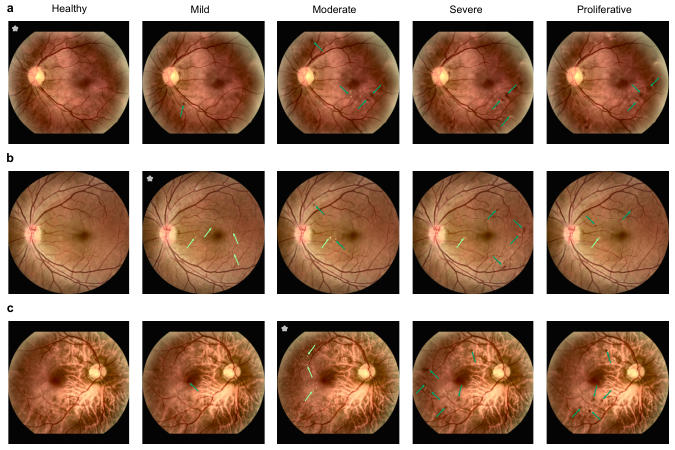}
\caption{DVCs for DR grading task with 5-classes: healthy, mild, moderate, severe and proliferative. Images marked with * are original images with GT as indicated in the headline. DVCs to the different classes from a \textbf{a.} healthy fundus, \textbf{b.} fundus with mild DR and \textbf{c.} fundus with moderate DR. Lesions which are originally present in initial image are indicated with light green arrows while lesions added by DVC are indicated with dark green arrows. In all cases, healthy DVCs removed all lesions. While the number and types of lesions introduced in mild and moderate DVCs are consistent with those observed in real-world data, severe and proliferative DVCs did not reflect the size and intensity of lesions seen in real examples.} 
\label{fig6}
\end{figure}

We followed up the counterfactuals for the binary case of healthy versus DR with counterfactuals for a more fine-grained classification scenario with $5$ classes, healthy, mild, moderate, severe and proliferative. The latter four categories are the various stages of DR in the order of increasing severity. The mild class often shows only very tiny changes in the form of microaneurysms and is the hardest to detect. Moderate and severe are characterized by the presence of a relatively greater number of microaneurysms and larger lesions such as hemorrhages and exudates. The proliferative class is the most advanced stage with venuous bleeding, large haemmorhages and neovascularization. Some of the images in the proliferative and severe stages also show scars resulting from laser treatment. Typically severe and proliferative classes are easier to detect due to larger lesions however due to rare occurrences they are underrepresented in the data set. 

We generated diffusion counterfactuals to the $5$ different classes from originally healthy, mild and moderate fundus images (Fig.\,\ref{fig6}). First, we looked at diffusion counterfactuals to the various DR stages from a healthy image and found that the diffusion counterfactuals contained meaningful features for both mild and moderate classes. The features included tiny dot-like microaneurysms/exudates for mild class and slightly larger and more exudates, microaneurysms and a few haemmorhages for moderate class (Fig.\,\ref{fig6}\,\textbf{a}). However, for the severe and proliferative classes most often only a couple of scattered haemmorhages were generated and most other features such as bleeding or the laser scars were not observed (Fig.\,\ref{fig6}\,\textbf{a}). This was likely due to the scarcity of these classes in the data set. Another technical factor could be the choice of parameters such as the regularization value $\lambda_d$ and the radius $\varepsilon$ which were more suitable for smaller changes. 

For healthy diffusion counterfactuals from both mild and moderate classes, all lesions were removed completely in most cases with a regularization strength of $0.5$ (Fig.\,\ref{fig6}\,\textbf{b-c}). On a moderate diffusion counterfactual generated from a fundus image originally belonging to the mild class, the number of exudates increased and an existing exudate was slightly enlarged (Fig.\,\ref{fig6}\,\textbf{b}). The diffusion counterfactual from moderate to mild interestingly removed the exudates all over the fundus and added a single microaneurysm (Fig.\,\ref{fig6}\,\textbf{c}). Here again, diffusion counterfactuals to the more advanced stages of severe and proliferative did not exhibit the relevant features for those classes (Fig.\,\ref{fig6}\,\textbf{b-c}).Furthermore, in all cases, the plain classifier achieved high target probabilities in the range $[0.85, 1.00]$ for diffusion counterfactuals to the healthy, mild and moderate classes while having only low target probabilities which dropped to below $0.25$ for the severe and proliferative diffusion counterfactuals. 

To summarize, the $5$-class model could generate meaningful diffusion counterfactuals to the healthy, mild and moderate classes while it was not as efficient at generating severe and proliferative cases. Nonetheless, mild and moderate classes are the clinically more interesting stages as they are challenging to detect and diagnostic decisions are uncertain for not only DNNs but also ophthalmologists around the boundaries of these early stages \cite{ayhan2020expert}. Studying the progression of biomarkers closely in these stages with counterfactuals can help prevent conversion to the more advanced stages.    

\subsection{Diffusion counterfactuals of OCT scans are also realistic}

Finally, we trained another set of diffusion model and classifiers on a database of $96,441$ OCT scans, a different image modality which is also predominantly used in ophthalmology (see Table \ref{tab:datasets}). While the diffusion model was trained to generate realistic OCT scans, the task of the classifiers was to detect whether a given scan was healthy or had one among the three conditions: choroidal neovascularization (CNV), drusen or Diabetic Macular Edema (DME), which both plain and robust classifiers were able to do with high accuracy (see Table \ref{tab:accuracies}). 

OCT scans can visualize a cross-section of the retina and typically show the different layers of the retina, from the vitreo-retinal interface, inner retina, outer retina, retinal pigment epithelium (RPE)/Bruch's membrane to the choroid from top to bottom. The biomarkers of CNV on OCT scans include subretinal neovascular membrane, subretinal fluid and intra-retinal fluid. These occur due to abnormal growth of new vessels in the choroid creating a rupture in the retinal layers above the Bruch's membrane. Drusen are characterized by a bumpy or irregular RPE layer due to lumps of deposits under the RPE. OCT scans of subjects with DME contain several cavity-like structures in the inner and sub retinal layers which represent intraretinal and subretinal fluid that accumulates due to vascular leakage \cite{oct_book}. Upon visual inspection, we found that the diffusion counterfactuals seemed to effectively capture salient features of the various classes. 

\begin{figure}[t]
\includegraphics[width=\textwidth]{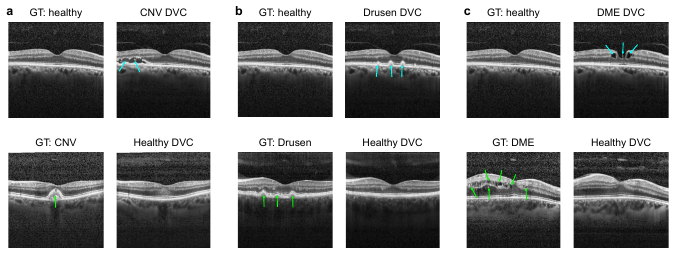}
\caption{DVCs of OCT images from healthy to various disease classes and vice-versa. \textbf{a.} DVC from healthy to CNV (top) and from CNV to healthy (bottom). \textbf{b-c.} Same as \textbf{a} for classes drusen (\textbf{b}) and DME (\textbf{c}). Similar to fundus DVCs, OCT DVCs show meaningful changes which are consistent with the important features of each class. DVCs from healthy images add features relevant to the disease (blue arrows). DVCs from diseased images to the healthy class remove the disease specific features seen on original image (green arrows).} 
\label{fig7}
\end{figure}

\begin{figure}[t!]
\includegraphics[width=\textwidth]{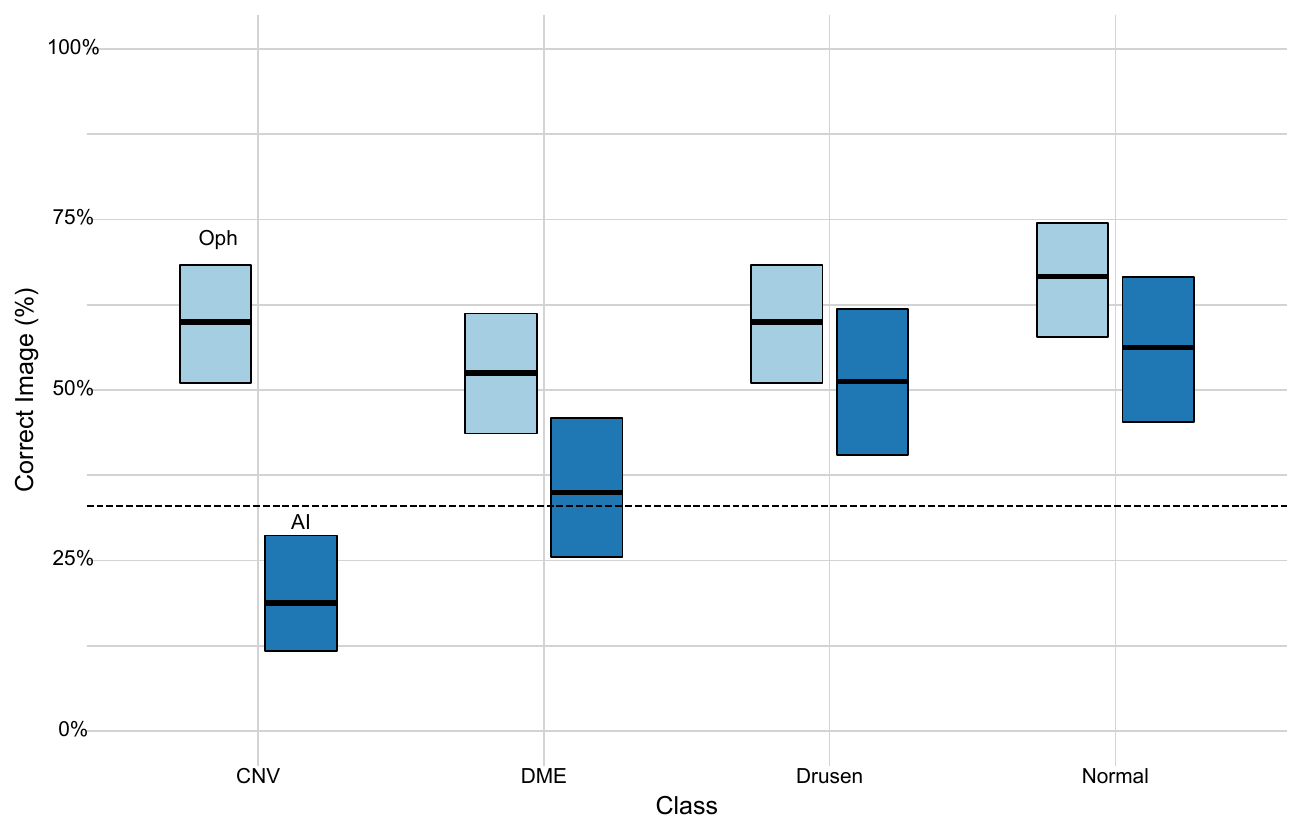}
\caption{Clinical evaluation of realism of generated OCT DVCs. We asked $n=4$ AI experts and $n=6$ ophthalmologists to identify a DVC in a odd-one-out task with three images (two real and once DVC). \textbf{a.} Overall fraction of correctly identified DVCs with binomial 95\%-CI. Baseline at $33\%$ (dashed line). \textbf{b.} As in \textbf{a.} } 
\label{fig8}
\end{figure}

We generated diffusion counterfactuals using the cone projection method with the OCT scans to the three disease categories from healthy and vice-versa. Diffusion counterfactuals contained meaningful and superficially realistic changes similar to fundus images. Diffusion counterfactual to CNV from healthy added subretinal fluid below the RPE (Fig.\,\ref{fig7}\,\textbf{a}, top row). On the other hand, the diffusion counterfactual from CNV to healthy removed the subretinal neovascular membrane and flattened out the portion where it was present (Fig.\,\ref{fig7}\,\textbf{a}, bottom row). While diffusion counterfactual from healthy to drusen class added bumps to the RPE layer(Fig.\,\ref{fig7}\,\textbf{b} top row), in the reverse case the irregularities were removed to make the RPE layer smooth and flat (Fig.\,\ref{fig7}\,\textbf{b} bottom row). DME diffusion counterfactuals generated from the healthy class contained cavities in the inner retinal layers (Fig.\,\ref{fig7}\,\textbf{c} top row). On the other hand, healthy diffusion counterfactual from a DME OCT scan covered up the cavities with the original tissue reflectivity (color) in those layers (Fig.\,\ref{fig7}\,\textbf{c} bottom row). Hence, in all cases, diffusion counterfactuals generated meaningful structures associated with the target class. 

To assess the degree of realism of the generated images, we performed a user study again with six ophthalmologists and four AI researchers. Similar to the fundus user study, the participants were assigned a three-way odd one out task here too although they were shown only diffusion counterfactuals across the four different categories. Ophthalmologists consistently performed better than chance ($33.3\%$) in all classes (Fig.\,\ref{fig8}, indicated by non-overlapping 95\%-CIs). Interestingly, they detected diffusion counterfactuals to the normal class from the various disease classes with the highest rate of $66.7\%$. This could have been due to  the normal diffusion counterfactuals generated from OCT scans with signs of extreme CNV or DME. In such scenarios, the normal DVCEs generally tended to fill up the cavities or tears with original tissue reflectivities but did not restore the thickness of the layers (Fig.\,\ref{fig7}\,\textbf{c} bottom row) thereby resulting in easier detection. They found diffusion counterfactuals to the CNV and drusen classes also easier to detect (Fig.\,\ref{fig8}). This could be due to certain features that looked artificially generated for e.g. the perfect waves on drusen diffusion counterfactuals. On the other hand, diffusion counterfactuals to the DME class was the hardest to detect for the ophthalmologists (Fig.\,\ref{fig8}). AI experts performed significantly worse overall (see Table \ref{tab:oct}) compared to ophthalmologists, this was especially true in the CNV and DME classes (Fig.\,\ref{fig8}). We attribute this to the relatively low experience of AI experts with these disease categories such that realistic looking images not showing realistic disease features led to performance at chance level for AI experts. 

\begin{table}
\centering
 \caption{Generalized Linear Model to assess the influence of factors in Fig. \ref{fig8}. $n=800$}
\begin{tabular}{cccc} 
 \hline
 Predictor & Odds Ratio & CI & p-value \\ 
 \hline
 CNV vs. DME & 1.09 & 0.73 - 1.63 & 0.6817 \\ 
 CNV vs. drusen & 1.72 & 1.15-2.58 & 0.0082 \\ 
 CNV vs. normal & 2.23 & 1.49 – 3.37 & 0.0001 \\ 
 Ophthalmologist vs. AI researcher & 0.44 & 0.33 - 0.60 & $< 0.0001$ \\ 
 \hline
 \label{tab:oct}
\end{tabular}
\end{table}

Taken together, OCT diffusion counterfactuals were able to generate the primary features associated with each class convincingly although with a few imperfections which led to their easier detection in the user study compared to fundus diffusion counterfactuals. 

\section{Discussion}

In conclusion, we showed that diffusion models guided by the gradients of robust classifiers can be used to generate realistic counterfactuals for retinal fundus and OCT images. We found that domain experts in ophthalmology including clinicians and AI researchers could hardly distinguish these images from real images, opening up new opportunities to include counterfactual images in medical reasoning \cite{prosperi2020causal}. 

While the counterfactuals for retinal fundus images were nearly indistinguishable from real ones for clinicians, OCT counterfactuals were relatively easier to detect. It is interesting to speculate what could have caused these differences. Likely, one factor is that lesions for early DR stages visible in fundus images are mostly localized and not too large, therefore not requiring major structural changes to large parts of the image, in contrast to what is needed for generating OCT counterfactuals. Also, OCT counterfactuals typically looked too regular and symmetric, such as in the case of drusen counterfactuals. Qualitative feedback after the user study indicated that raters were  quickly able to pick up on these regularities. In addition, healthy OCT counterfactuals from extremely diseased cases often covered the abnormalities with appropriate texture but did not alter the thickness of the retina at that stage of the image which is an important factor for clinicians to classify the image as healthy. Since most cases in the chosen OCT data set also belong to extreme disease stages, this could have further impacted the overall performance of clinicians in the detection of OCT counterfactuals. In contrast, fundus images covered the whole disease spectrum, allowing the model to learn about the gradual changes along the disease trajectory.

It is possible that the different degrees of realism of fundus vs. OCT counterfactuals come either from the generative capabilities of the diffusion model or  what is learned by the classifiers. In fact, we noticed that supplementing the fundus image dataset with more examples from diseased classes helped to generate better disease counterfactuals. However, it is also possible that the disease concepts learned by the classifiers which guide the diffusion model are insufficient. For example, with the OCT dataset consisting mostly of extreme examples at advanced disease stages and missing much of the borderline cases in between \cite{kermany_2018}, a classifier might have easily taken a short-cut towards distinguishing healthy from diseased images, picking up on the most informative feature (e.g. texture) while ignoring more subtle ones (retinal thickness) \cite{geirhos2020shortcut}. In turn, improved classifiers based on more realistic and varied datasets may therefore yield even better counterfactuals. In fact, it would be interesting to study how the generative capacities of the model change with even larger datasets, such as those used to train retinal foundation models, or based on robust classifiers derived from such foundation models \cite{zhou2023foundation}.

Diffusion models have been largely used in combination with plain classifiers for realistic counterfactual generation in the natural image domain using data sets such as ImageNet and CelebA \cite{diffusionCEs_2022, dime_2022}. The quality and realistic nature of counterfactuals for such images has been shown to improve when adversarially robust classifiers are used \cite{dvce_neurips2022}. In the medical setting, diffusion models have been used primarily for generating healthy counterfactuals \cite{anomaly_cfs_2022, anomaly_diff_models_2023, lesion_cfs_2022}, which is an easier task for the diffusion model compared to generating disease related features. We demonstrated that both diffusion models and adversarially robust classifiers play a major role in generating realistic medical counterfactuals for high-resolution retinal fundus and OCT images. Moreover, our counterfactuals are bi-directional, i.e. from healthy to diseased and diseased to healthy. In parallel work, BioMedJourney \cite{biomedjourney_arxiv2023} uses two consecutive Chest-XRay images of a subject and a summary of their medical reports to generate longitudinal counterfactuals. Here, a latent diffusion model is trained on embeddings of the textual descriptions and a starting image to obtain an estimate of the progressing image. This method relies on the availability of detailed medical reports in addition to longitudinal imaging data. 
 
Realistic counterfactuals have the potential to be used in clinical decision support where the DNN can provide human-like and human-understandable reasoning for it's prediction. For example, decision support is conceivable that illustrates for a given patient with an uncertain diagnosis how the imaging data might look if it provided less ambiguous evidence for the presence of a disease, potentially even with more than one sample for a given input image. The clinician could then use similarity of the present image to judge the presence or absence of disease signs.  Realistic counterfactual images could also be used to synthesize data for training clinicians and augmenting DNN models, as medical data sets are often imbalanced and diseased samples may be less readily available. Due to the realistic nature of counterfactuals, diseased data points could then be synthesized based on a preliminary classifier from the more prevalent healthy examples \cite{synthetic_LDMs_miccai2023, synthetic_DMs_neurips2022ws}. In fact, adding a diseased counterfactual for each healthy image and vice versa would effectively create a paired dataset, where structurally similar images derived from the same base image are contained in the healthy and diseased class, allowing the classifier to focus more easily on the disease patterns. In the reverse case of generating healthy examples from diseased, the counterfactuals could help in anomaly detection and identification of bio-markers \cite{anomaly_cfs_2022, anomaly_diff_models_2023, lesion_cfs_2022}. Further, they could be used as a testing tool to ensure that the classifiers do not use any shortcuts to make the decisions, such as hospital or device logos instead of disease related features. 

A natural extension of this work would be to generate counterfactuals from multi-task DNNs which learn several attributes simultaneously \cite{multitask_tvst2023}. With such a DNN, it would be possible to generate counterfactuals for one attribute keeping another fixed. For instance, a multi-task classifier which is trained on both age and disease type can be used to generate counterfactuals for increasing age keeping the disease fixed or vice-versa. Such counterfactuals could potentially be used in tracking the progression of a disease with age. Similarly, it would be interesting to study counterfactuals for longitudinal data or data with interventions, such as the administration of a drug. For example for OCT images during age-related macular degeneration, treatment effects for the injection of anti-VGEF drugs might be simulated for the different available drugs, and the most promising drug chosen. 

\subsection*{Acknowledgments}

We acknowledge support by the German Ministry of Science and Education (BMBF; 01IS18039A), the Deutsche Forschungsgemeinschaft through a Heisenberg Professorship (BE5601/8-1) and under Germany’s Excellence Strategy – Excellence Cluster "Machine Learning --- New Perspectives for Science" EXC2064/1 –-- Project number 390727645), the Carl Zeiss Foundation (project “Certification and Foundations of Safe Machine Learning Systems in Healthcare”) and Gemeinnützige Hertie Stiftung. This research utilized compute resources at the Tübingen Machine Learning Cloud, INST 37/1057-1 FUGG. PB is a member of the Else Kröner Medical Scientist Kolleg "ClinbrAIn: Artificial Intelligence for Clinical Brain Research”. 

\pagebreak 
\section{Appendix}

\begin{figure}[h!]
\includegraphics[width=\textwidth]{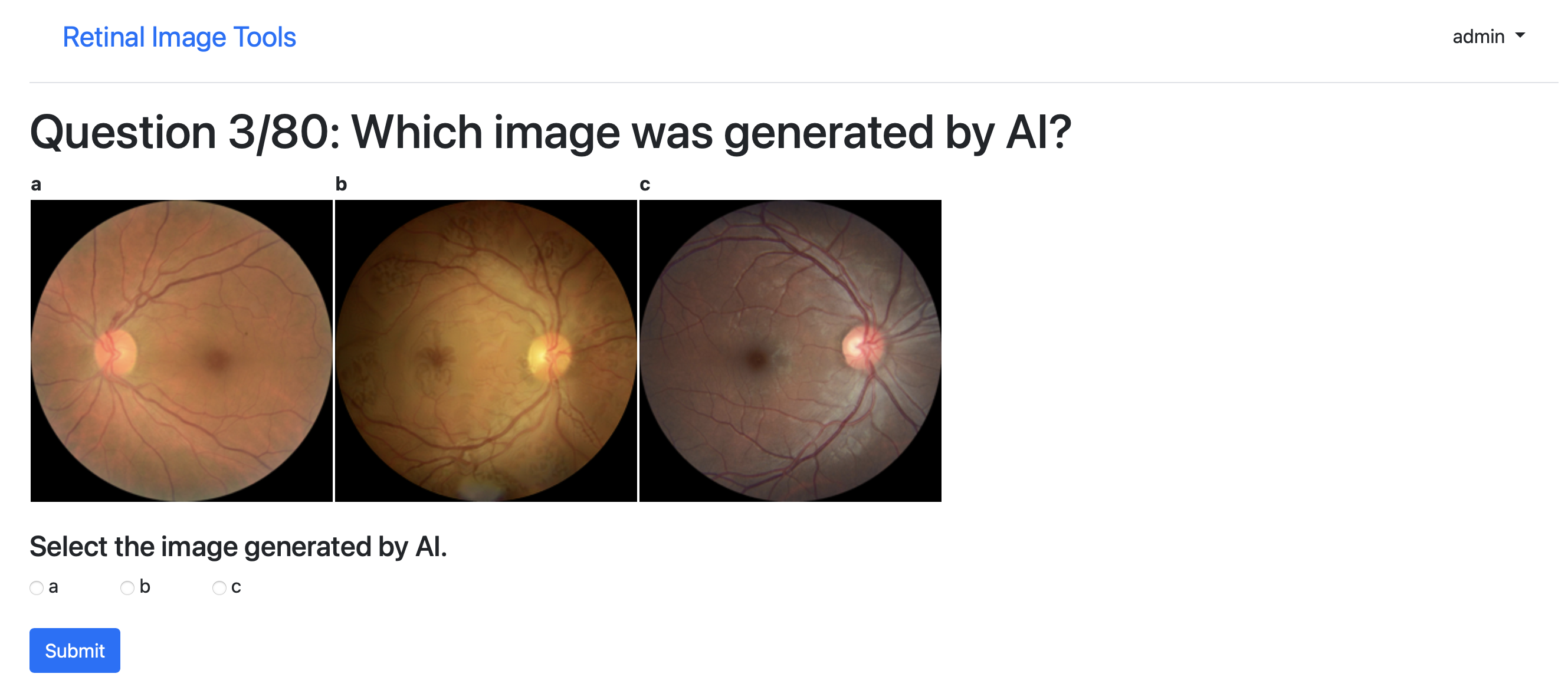}
\caption{Web interface for evaluating realism of counterfactuals. Three images are shown on the page where two are real and one is generated. User is asked to select the generated image}
\label{app1}
\end{figure}

\begin{figure}[h!]
\includegraphics[width=\textwidth]{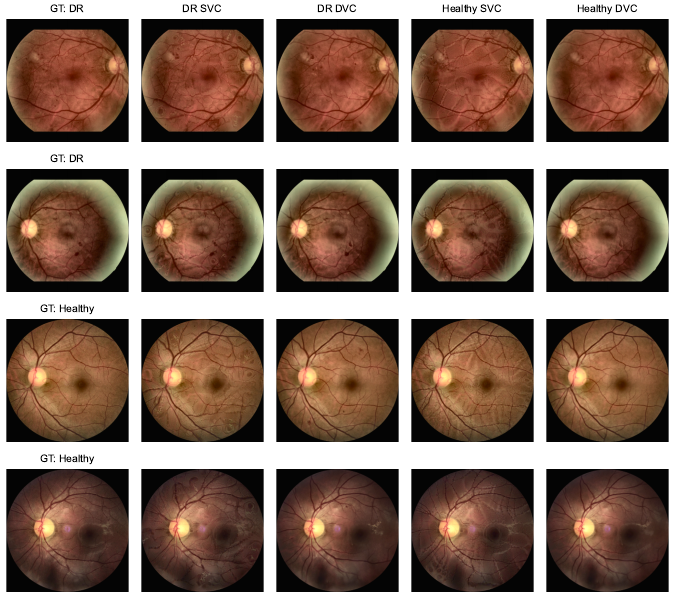}
\caption{More examples comparing SVCs and DVCs. Top two rows show counterfactuals from DR fundus images. Bottom two rows show counterfactuals from healthy images. In all cases, changes in DVCs are more realistic compared to SVCs.}
\label{app2}
\end{figure}

\begin{figure}[h!]
\includegraphics[width=\textwidth]{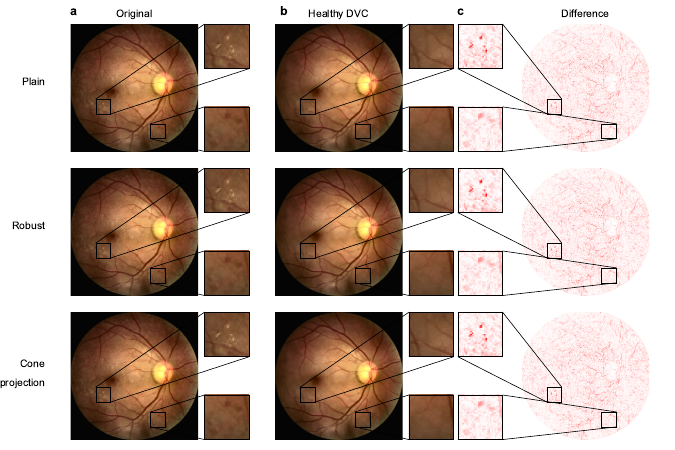}
\caption{As in Fig.\,\ref{fig4} for a DR fundus image. \textbf{a}. Original image with ground truth label DR. \textbf{b}. DVC to the healthy class. Plain model (top row) removes lesions to a similar extent as robust (middle row) and cone projection (bottom row). DVCs to healthy class are more easily generated than to DR class. \textbf{c} Difference maps between the original DR image and generated healthy counterfactual highlighting lesion locations.}
\label{app3}
\end{figure}

\begin{figure}[h!]
\includegraphics[width=\textwidth]{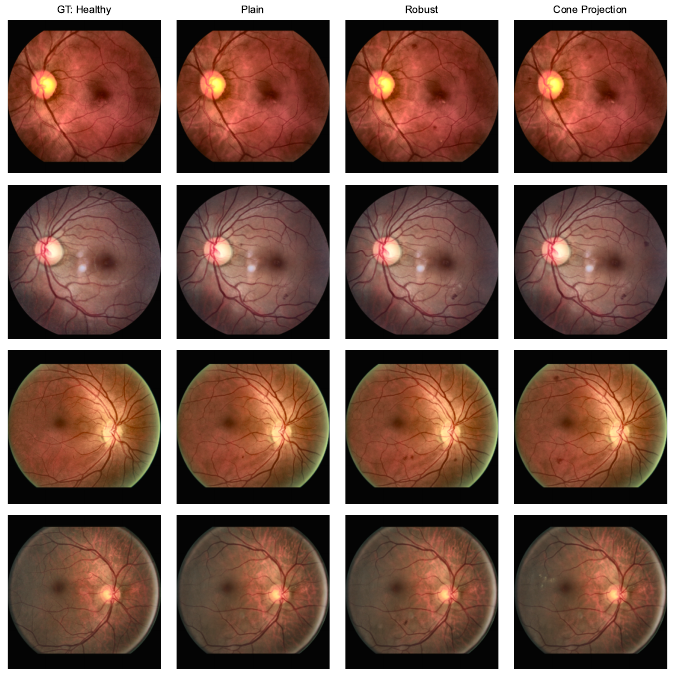}
\caption{More examples of DR DVCs generated from healthy fundus images (leftmost column) using the plain model gradients (second column), robust model gradients (third column) and cone projected gradients (rightmost column). In all examples, plain models either show no or fewer and weaker lesions compared to robust and cone projection models.}
\label{app4}
\end{figure}

\begin{figure}[h!]
\includegraphics[width=\textwidth]{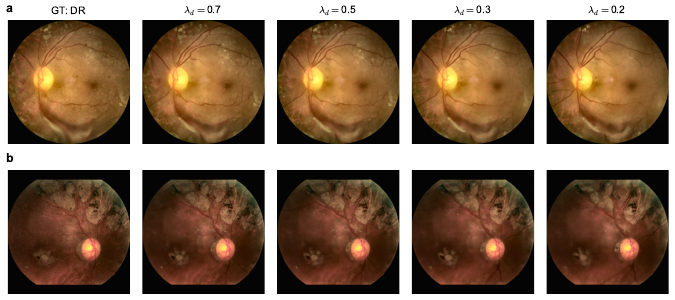}
\caption{Effect of regularization strength on retinal fundus images severely affected by DR. In such extreme cases, even a small regularization of $0.2$ is not sufficient to convert the image to healthy.}
\label{app5}
\end{figure}

\begin{figure}[h!]
\includegraphics[width=\textwidth]{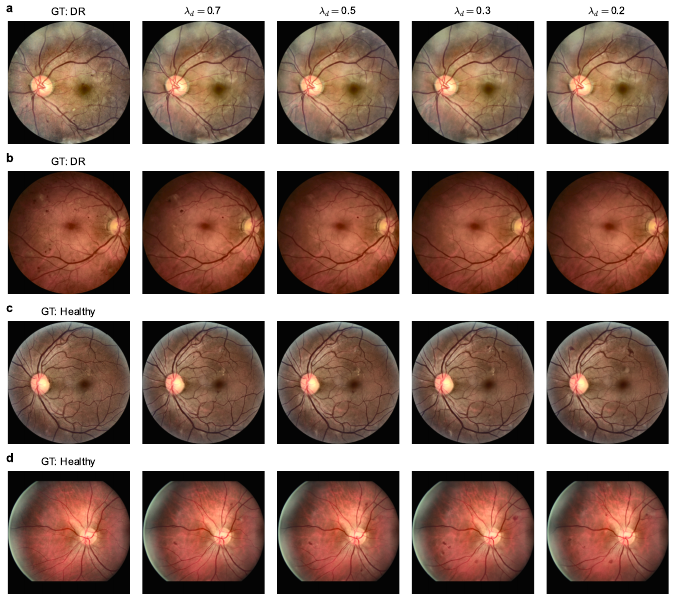}
\caption{Effect of regularization strength on retinal fundus images belonging to DR and healthy class. \textbf{a-b}. Healthy DVCs of DR fundus images with different values of $\lambda_d$. With $\lambda_d=0.5$, examples are converted to the healthy class with either no lesions (\textbf{a}) or very few remaining lesions (\textbf{b}) such as in the mild DR class. \textbf{c-d} DR DVCs of healthy fundus images with varying $\lambda_d$. Here too, with $\lambda_d=0.5$, the DVC adds enough lesions to change the decision of the classifier to the DR class with high confidence.}
\label{app6}
\end{figure}

\bibliographystyle{unsrt}
\bibliography{refs}

\end{document}